\documentclass[10pt,twocolumn,letterpaper]{article}

\usepackage{cvpr}              

\usepackage{makecell} 
\usepackage{booktabs}        
\usepackage{graphicx}        
\usepackage{array}           
\usepackage{adjustbox}       

\usepackage{xcolor}
\usepackage{xspace}
\usepackage{multirow}
\usepackage{pifont}
\usepackage[table]{xcolor}

\def\dataset{RealAppliance\xspace}
\def\benchmark{RealAppliance-Bench\xspace}

\definecolor{darkergreen}{RGB}{19,168,33}
\newcommand{\greencheck}{{\color{darkergreen}\ding{51}}}
\newcommand{\redcross}{{\color{red}\ding{55}}}

\definecolor{cvprblue}{rgb}{0.21,0.49,0.74}
\usepackage[pagebackref,breaklinks,colorlinks,allcolors=cvprblue]{hyperref}

\def\paperID{16962} 
\def\confName{CVPR}
\def\confYear{2026}

\title{\dataset: Let High-fidelity Appliance Assets \\ Controllable and Workable as Aligned Real Manuals}

\author{
Yuzheng Gao\textsuperscript{*1},
Yuxing Long\textsuperscript{*1,2},
Lei Kang\textsuperscript{1,2},
Yuchong Guo\textsuperscript{1},
Ziyan Yu\textsuperscript{1},
Shangqing Mao\textsuperscript{1}, \\
Jiyao Zhang\textsuperscript{1,2}, 
Ruihai Wu\textsuperscript{1},
Dongjiang Li\textsuperscript{2},
Hui Shen\textsuperscript{2} and
Hao Dong\textsuperscript{\dag1}\\
\textsuperscript{\rm1}CFCS, School of Computer Science, Peking University\\
\textsuperscript{\rm2}Jingdong Technology Information Technology Co., Ltd\\
*Equal contribution, \dag~Corresponding author\\
https://realappliance.github.io/
}

\begin{document}

\twocolumn[{%
\renewcommand\twocolumn[1][]{#1}%
\maketitle
\begin{center}
    \centering 
    \vspace{-2em}
    \includegraphics[width=0.9\linewidth]{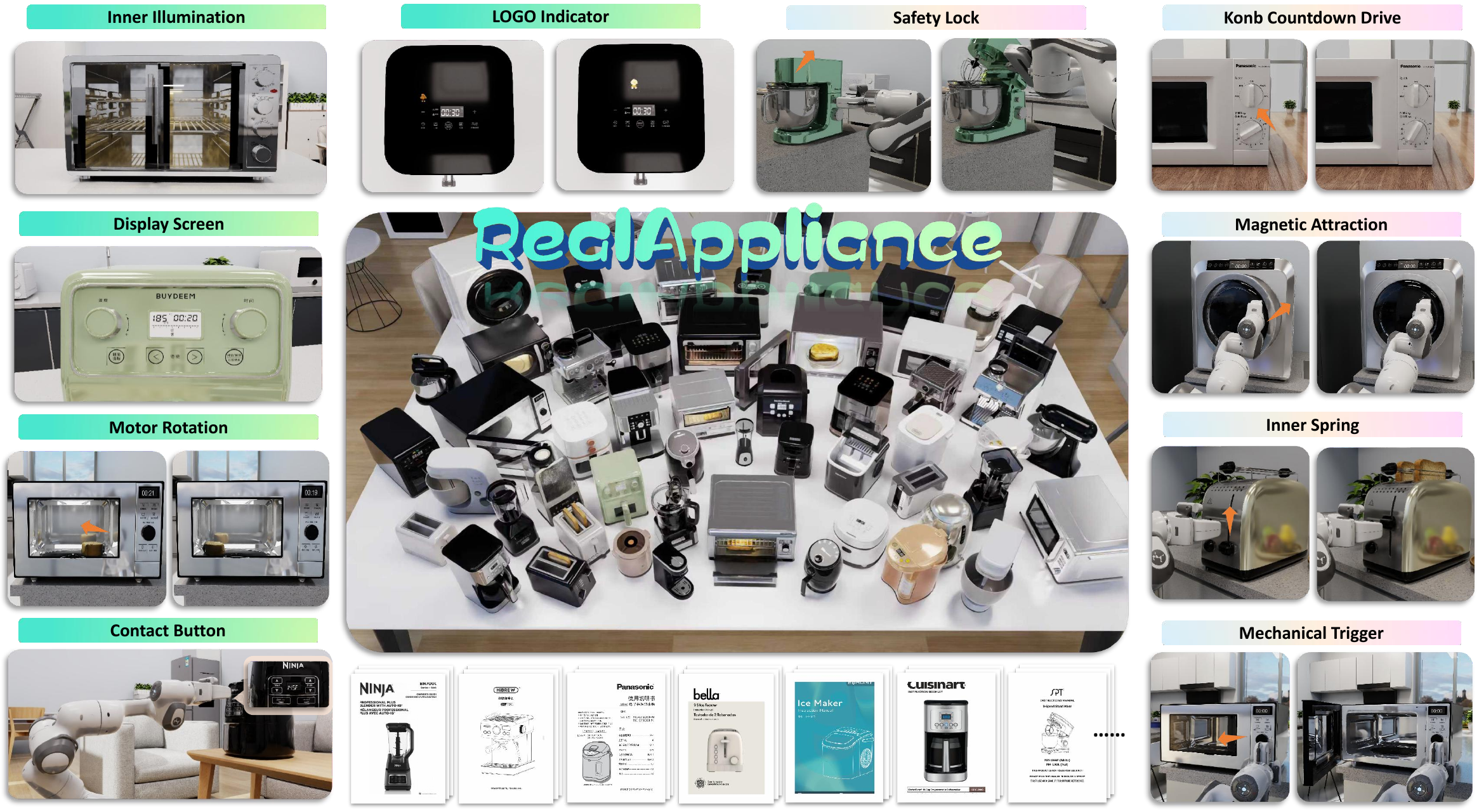}
    \captionof{figure}{\textbf{Overview of \dataset.} We collect 100 real appliance manuals and create 100 high-fidelity appliances digital assets aligned with these manuals. Every appliance asset has the same size, texture and physical mechanisms, electronic mechanisms as the real one. }
    \label{fig:teasor}
\end{center}
 }]

\begin{abstract}
Existing appliance assets suffer from poor rendering, incomplete mechanisms, and misalignment with manuals, leading to simulation-reality gaps that hinder appliance manipulation development. In this work, we introduce the \dataset dataset, comprising 100 high-fidelity appliances with complete physical, electronic mechanisms, and program logic aligned with their manuals. Based on these assets, we propose the \benchmark benchmark, which evaluates multimodal large language models and embodied manipulation planning models across key tasks in appliance manipulation planning: manual page retrieval, appliance part grounding, open-loop manipulation planning, and closed-loop planning adjustment. Our analysis of model performances on \benchmark provides insights for advancing appliance manipulation research.

\end{abstract}    
\section{Introduction}
\label{sec:intro}
Home appliances are common in our daily life. Mastering appliance operation is crucial for future home service robots to provide autonomous assistance. To conduct effective research on appliance manipulation planning, it is important to build high-fidelity appliance assets that closely resemble real appliances. Such appliance assets should achieve high realism across multiple dimensions:
First, in appearance, models should replicate real-world geometric dimensions, surface textures, and materials as closely as possible.
Second, in functionality, all movable components reproduce physical and electronic mechanisms, and the program logic between components is also consistent with that of real appliances.
Additionally, to enable robots to understand operational procedures, each appliance should include aligned real manuals.

However, existing commonly used appliance assets fall short in these aspects. For instance, the models in PartNet-Mobility~\cite{sapien} exhibit low rendering quality and mechanism-free components. CheckManual~\cite{checkmanual} attempts to generate manuals based on these assets, yet its textual descriptions and illustrations remain far from those in real manuals. ArtVIP~\cite{artvip} considers certain appliance functionalities but provides only a limited set of assets, and some components such as knobs are non-operable. These limitations hinder the development of appliance manipulation planning.

To address these issues, we introduce a new appliance asset dataset, \dataset. In the modeling phase, we collect manuals from various countries and regions, using their dimensional data and real photos to create detailed 3D models of appliances. All externally visible parts are modeled independently with precise collision bodies. For textures, we perform UV unwrapping and design high-resolution UV maps, referencing real photos and manual details (\emph{e.g.}, logos and scales) to faithfully reproduce the visual and structural characteristics of real appliances. 

To achieve functional fidelity, we configure both physical and electronic mechanisms for each appliance model following the appliance manual. Specifically, we assign appropriate joint types and physical parameters to each movable part and implement two key categories of mechanisms:
(1) Physical mechanism (\emph{i.e.}, inner spring, magnetic attraction, knob countdown, and safety lock); and
(2) Electronic mechanism (\emph{i.e.}, screen display, touch sensing, illumination, logo indicator, and rotary motor)
Furthermore, we design the program logic of appliances based on their manuals. For example, pressing a touch button changes screen content, starts mixer rotation, or toggles indicator lights, thereby reproducing real operational workflows in simulation.

With above assets, we introduce \benchmark, a comprehensive benchmark for evaluating models' capabilities in appliance manipulation planning. It encompasses four tasks: manual page retrieval, appliance part grounding, open-loop manipulation planning, and closed-loop planning adjustment. These tasks span the full pipeline from manual understanding and action planning to feedback correction, requiring skills in long-document reasoning, fine-grained part localization, long-horizon planning, and online error adjustment. We systematically assess leading multimodal large models and embodied planning models on \benchmark, offering quantitative and qualitative analyses to highlight common challenges and guide future research.

In this work, our main contribution includes:
\begin{itemize}
    \item[$\bullet$] We present \dataset, the first asset dataset fully constructed in accordance with manuals on vision and function. Every asset in \dataset comes with a corresponding real manual.
    \item[$\bullet$] We design manual-aligned mechanisms and program logic for each appliance asset in \dataset, making the assets controllable and functional in simulation just like real appliances.
    \item[$\bullet$] We introduce the \benchmark benchmark, which features key tasks in appliance manipulation planning. We comprehensively evaluate state-of-the-art multimodal large models and embodied planning models to highlight common challenges and guide future research.
\end{itemize}
\section{Related Work}
\begingroup
\setlength{\tabcolsep}{4pt} 
\renewcommand{\arraystretch}{1.2} 
\begin{table*}[htbp]
\centering
\caption{\textbf{Comparison with existing appliance digital assets.}}
\scalebox{0.8}{%
\begin{tabular}{@{}l|*{9}{c}@{}} 
\toprule
\makecell{Digital Assets} & 
\makecell{Appliance\\Category} & 
\makecell{Appliance\\Number} & 
\makecell{Real\\Size} &
\makecell{Real\\Texture} &
\makecell{Movable\\Joints} & 
\makecell{Physical\\Logic} & 
\makecell{Electronic\\Components} &
\makecell{Electronic\\Logic} & 
\makecell{Appliance\\Manual} \\
\midrule
Partnet-Mobility~\cite{sapien} & 17 & 636 & \redcross & \redcross & \greencheck & \redcross & \redcross & \redcross & \redcross \\
CheckManual~\cite{checkmanual} & 11 & 369 & \redcross & \redcross & \greencheck & \redcross & \redcross & \redcross & Synthesis Manuals \\
Infinite Mobility~\cite{lian2025infinite} & 5 & -- & \redcross & \redcross & \greencheck & \redcross & \redcross & \redcross & \redcross \\
ArtVIP~\cite{artvip} & 12 & 49 & \greencheck & \greencheck & \greencheck & \greencheck & \redcross & \redcross & \redcross \\
\dataset\ (Ours) & 14 & 100 & \greencheck & \greencheck & \greencheck & \greencheck & \greencheck & \greencheck & Real Manuals \\
\bottomrule
\end{tabular}
}
\end{table*}
\endgroup

\subsection{Appliance Digital Assets}
Household appliances, as common objects in daily life, have long been important targets for digital asset modeling.  Partnet-Mobility~\cite{sapien} treats appliances as articulated objects, configuring different types of joints for movable components such as knobs, buttons, and doors. However, these models neither align with real appliances in terms of dimensions and textures nor provide accompanying manuals. Infinite Mobility~\cite{lian2025infinite} focuses on scaling up the quantity of appliance models by designing automated generation algorithms capable of producing unlimited assets for a limited set of categories. Nevertheless, this approach still fails to address issues related to the realism and functionality of the assets. CheckManual~\cite{checkmanual} recognizes the necessity of appliance manuals for proper operation and combines automatic generation with manual verification to produce manuals for assets from Partnet-Mobility. Yet, due to limitations in both the asset models and the manual generation algorithms, a significant gap remains between these and real appliances or manuals. ArtVIP~\cite{artvip} addresses the authenticity of interaction logic for articulated assets, including appliances, by incorporating damping, magnetic attraction, and trigger mechanisms into the modeling process. However, it still does not implement the operational procedures of appliances. Distinct from the aforementioned works, our proposed \dataset is the first to model appliance assets directly based on real manuals, ensuring alignment in dimensions, textures, mechanisms, and program logic with the original manual, thereby introducing new challenges and opportunities for research on realistic appliance operation.

\subsection{Manual-based Appliance Manipulation}
Operating home appliances differs from using passive tools, as it requires handling each component in the correct sequence and manner to ensure proper functionality. Appliance manuals typically contain prior information such as component names and operating procedures, yet previous research~\cite{li2024manipllm}\cite{openvla}~\cite{voxposer}~\cite{eisner2024flowbot3dlearning3darticulation}~\cite{black2024pi_0} all bypasses manuals, instead relying on large models to plan actions from commonsense knowledge. Such knowledge is rarely aligned with specific appliances, limiting operations to short-range interactions. In recent years, researchers have begun to recognize the value of manuals for appliance operation. ApBot~\cite{apbot} and ManualPlan~\cite{checkmanual} plan manipulation based on manual content. Nevertheless, their evaluation methods have clear shortcomings: ApBot's state machine-based evaluation lacks visual feedback and assumes direct access to accurate post-operation state information, which is unrealistic for real scenarios. ManualPlan relies on synthetic manuals whose text and images differ significantly from real manuals. To address these issues, we developed \benchmark, a benchmark based on real manuals and controllable, operable appliance assets, providing realistic visual feedback during operations. This enables comprehensive and effective evaluation of a model’s capabilities in manual understanding, component grounding, action planning, and closed-loop adjustment.  

\begin{figure*}[htp]
    \centering
    \includegraphics[width=0.9\linewidth]{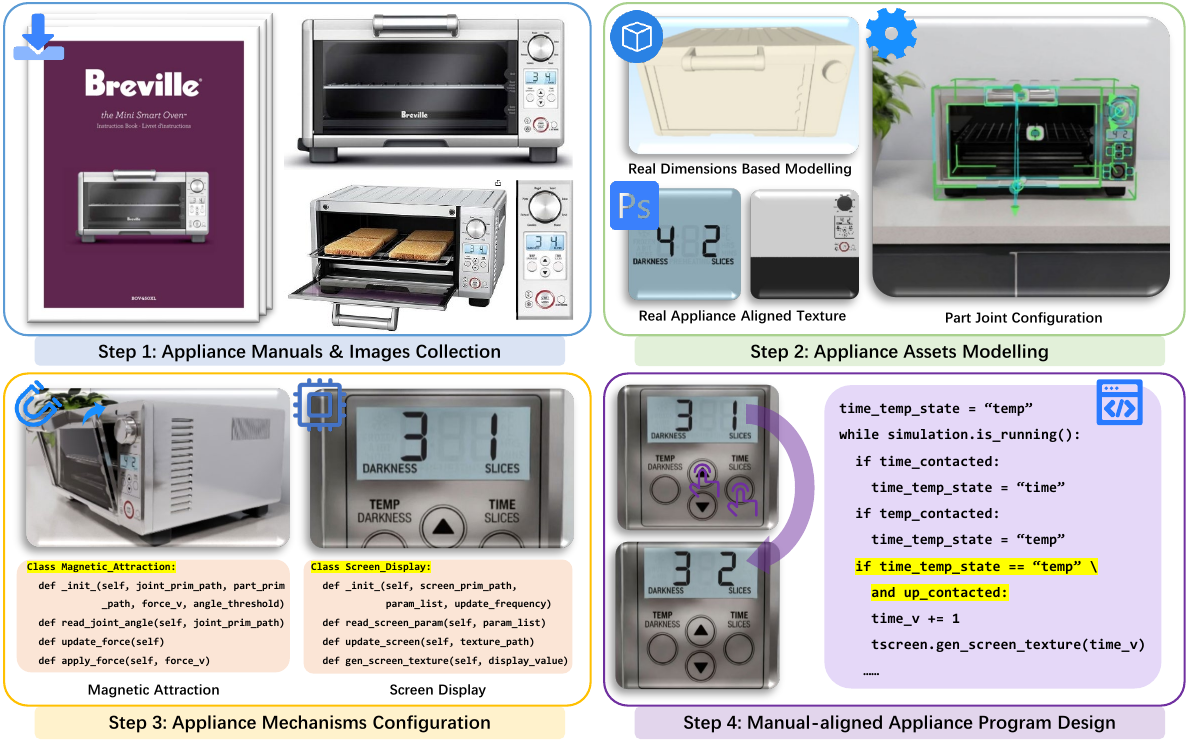}
    \vspace{-0.2cm}
    \caption{\textbf{Creation process of appliance digital assets in \dataset.}}
    \label{fig:data_creation}
    \vspace{-0.3cm}
\end{figure*}

\section{\dataset Digital Assets}
\subsection{Appliance Manuals and Photos Collection}
We collected household appliances and their corresponding user manuals from multiple countries, following four principles: (1) Excluded appliances with buttons too small for robotic operation; (2) Selected manuals of moderate length to fit MLLM capacity; (3) Ensured clear descriptions of components and operating procedures; (4) Required manuals to include dimensions and high-resolution product photos.
Under these criteria, we obtained 100 manuals covering 14 appliance categories in languages such as Chinese, Russian, French, German, and others. For each appliance, multi-angle photographs were taken to capture logos, scale markings, and surface materials, along with measurements of actual dimensions.

\subsection{Appliance Digital Asset Creation}

\noindent\textbf{Appliance Modeling}
We model each appliance in Autodesk 3Ds Max~\cite{3dsmax} based on manuals, photos, and actual sizes.  Functional parts are modeled as separate components, and TurboSmooth is applied to increase polygon density and enhance visual quality. UV mapping is carefully unwrapped and refined in Unfold3D~\cite{unfold3d}. Screen and touch-control areas are isolated to allow easy texture updates. Based on these UV layouts, color textures are produced in Adobe Photoshop~\cite{photoshop}, faithfully replicating surface colors, proportions, icons, and logos from reference materials to ensure high visual fidelity.  

\noindent\textbf{Asset Setup}
We import the assembled appliance models and color textures into the NVIDIA Isaac Sim~\cite{isaac} platform to generate USD-format digital assets. Each appliance is defined in a right-handed coordinate system, with the geometric center as the origin: the z-axis points upward, and the y-axis faces the front of the appliance. Components are named according to the terminology in the corresponding user manual to facilitate easy retrieval. Material properties are adjusted to accurately reproduce glass, plastic, and metal surface effects on the appliances.

\noindent\textbf{Joint Design}
We then configure joint parameters for appliance components in Isaac Sim, defining positions, types, and motion limits. Rotational joints are used for elements such as knobs, hinged doors, and flip lids that require pivoting movement. Prismatic joints are applied to components like mechanical buttons, sliders, and push–pull doors that move linearly. Fixed joints are assigned to touch buttons, screen, and other non‑movable interfaces.

\subsection{Appliance Mechanisms Configuration}
To enable appliance assets to respond to interactions as real appliances do, we have designed and implemented a set of mechanisms. Each mechanism is encapsulated in an independent class that follows a unified interface specification, allowing for modular combination, replacement, or extension. This object‑oriented architecture enables flexible configuration of functional modules for appliance assets, delivering an interactive simulation experience that closely mirrors real‑world appliances. These mechanisms are broadly classified into physical mechanisms and electronic mechanisms, which are introduced in detail in the following.

\noindent\textbf{Physical Mechanisms}
These mechanisms replicate tangible, force‑driven behaviors found in real appliances, enabling components to interact through mechanical motion, pressure, or magnetic effects. These elements simulate the structural and kinetic operations that occur without relying on electronics.
\begin{itemize}
    \item[$\bullet$] \textbf{\textit{Inner Spring}}: Applies force from a compressed or stretched spring to return a component to its original position or assist movement. For example, a toaster lever uses spring force to pop up toast and reset after the cycle.
    
    \item[$\bullet$] \textbf{\textit{Magnetic Attraction}}: Uses magnetic force to hold components together or ensure secure closure. For example, washing machine doors employ magnetic strips for a tight seal during working to maintain temperature.
    
    \item[$\bullet$] \textbf{\textit{Mechanical Trigger}}: Achieves mechanical trigger relationships among components via causal logic. For example, pressing a microwave's door-open button pops the door; pressing a close button resets all pressed buttons.

    \item[$\bullet$] \textbf{\textit{Knob Countdown Drive}} Produces a countdown effect by mechanically rotating a knob. For example, an air fryer's timer knob rotates back to zero during working, stopping the appliance upon reaching it.
    
    \item[$\bullet$] \textbf{\textit{Safety Lock}}: Prevents physical operation when locked. For example, a mixer's motor head safety lock requires pressing a button or turning a knob to allow lifting.
\end{itemize}

\noindent\textbf{Electronic Mechanisms}
These mechanisms reproduce sensor‑based, motor‑driven, and display‑oriented functionalities, allowing appliances to respond to inputs and update their states dynamically. They enable visual feedback, automated motion, and interactive controls, closely mirroring the electronic mechanisms present in modern appliances.
\begin{itemize}
    \item[$\bullet$] \textbf{\textit{Screen Display}}: Generates the texture of screen area in the real time to display the appliance's current states. For example, a smart oven can display current temperature and time settings on the screen.
    
    \item[$\bullet$] \textbf{\textit{Touch Sensing}}: Binds a virtual contact sensor to a touch button to detect external physical force trigger corresponding control actions. For example, a coffee machine with a touch button allows users to start or stop the brewing process by simply tapping the button area.
    
    \item[$\bullet$] \textbf{\textit{Illumination}}: Controls the interior lighting of appliances in response to state updates. For example, the light inside a microwave automatically switches on when the microwave begins working.
    
    \item[$\bullet$] \textbf{\textit{Logo Indicator}}: Updates indicator lights on the control panel in sync with appliance state changes to communicate high-level status information. For example, a washing machine flashes its status icon to signal completion.  
    
    \item[$\bullet$] \textbf{\textit{Rotary Motor}}: Drives the joints of an appliance's movable components to simulate motor operation. For example, the microwave turntable rotates slowly to ensure heating evenly.
\end{itemize}

\begin{figure*}[htp]
    \centering
    \includegraphics[width=0.9\linewidth]{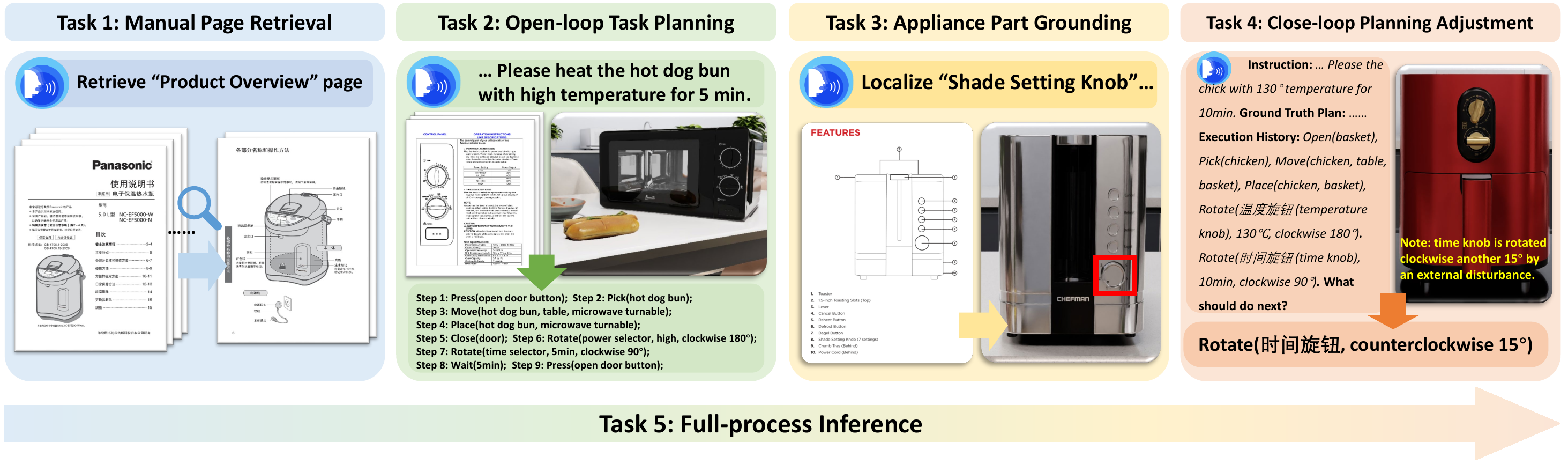}
    \vspace{-0.2cm}
    \caption{\textbf{Evaluation tasks in \benchmark.} These tasks cover the essential capabilities in appliance manipulation planning.}
    \label{fig:data_vis}
    \vspace{-0.3cm}
\end{figure*}

\subsection{Manual-aligned Appliance Programs}
With our designed physical and electronic mechanisms, we developed a program script for each appliance based on its corresponding manual to ensure that the appliance behaves in the simulation like a real one. The development process for these program scripts is detailed below.

\noindent\textbf{Determine Appliance Setting Parameter}
We define the setting parameters of appliances based on the content of the manual, such as power status, temperature, time, and working mode, and determine the candidate value ranges for these parameters (\emph{e.g.}, power status as a binary variable of 0 or 1). These setting parameters serve as the link for information transmission between appliance components, enabling updates to component states.

\noindent\textbf{Configure Appliance Part Mechanism}
We further configure the physical and electronic mechanisms for appliance parts according to the manual. Each mechanism class for every component inherits from the base class of that mechanism type, with parameters and functions modified to reflect the functional characteristics of the appliance. 

\noindent\textbf{Design Appliance Program Logic}
Once the setting parameters and mechanisms are established, we design the appliance’s program logic following the operational procedures described in the manual. The logic relies primarily on evaluating the states of the setting parameters and updating component states accordingly. When a parameter’s value enters a predefined range, the system updates the relevant component states and may also adjust other parameters as required.

\section{\benchmark}
\dataset contains real appliance manuals along with appliance assets aligned to these manuals, providing favorable conditions for evaluating models' ability in appliance manipulation planning. Based on the \dataset dataset, we propose \benchmark to assess a range of key capabilities for appliance manipulation planning with manuals.

\begin{figure*}[htp]
    \centering
    \includegraphics[width=\linewidth]{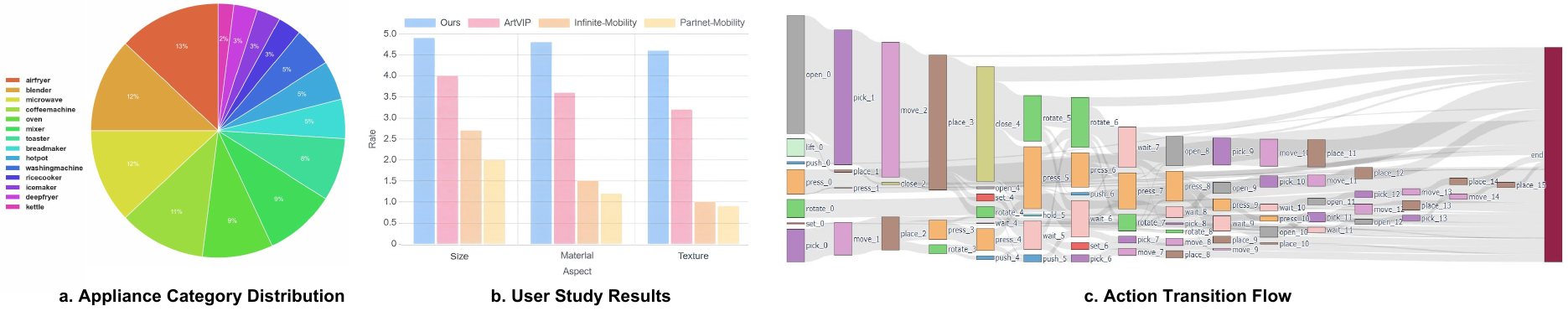}
    \vspace{-0.2cm}
    \caption{\textbf{Statistics visualization of \benchmark.} (Left) We display the proportion of various types of appliances via pie chart. (Middle) We conducted a user study about the fidelity of appliance asset(Right) We visualize the atomic action transitions in task planning to demonstrate planning diversity.}
    \label{fig:data_vis}
    \vspace{-0.3cm}
\end{figure*}

\subsection{Evaluation Tasks}
\noindent\textbf{Task 1: Manual Page Retrieval}  
Appliance manuals typically include component descriptions, operating procedures, safety precautions, routine maintenance, and after-sales service. For appliance manipulation planning, component descriptions and operating procedures are most relevant, while other sections are less related. To reduce inference overhead, we design a manual-page retrieval task: given a manual and a target page category, the model identifies relevant pages. Performance is evaluated using precision and recall.

\noindent\textbf{Task 2: Open-loop Task Planning}  
The manual serves as a key guide for appliance manipulation, introducing component names and operation procedures. The open-loop planning task aligns with real-world appliance manipulation scenarios, requiring the model to plan complete steps based on task instructions, the manual, and an initial observation image. To ensure task validity, we gathered 50 candidate tasks per appliance via a questionnaire and selected high-quality tasks for evaluation. To guarantee the executability and evaluability of the planning results, we defined 9 types of appliance manipulation actions (\emph{e.g.}, Press(part\_name, target\_state, press\_times), Rotate(part\_name, target\_state, rotate\_degrees), Open/Close(target\_part),) and 4 types of atomic object manipulation actions (\emph{e.g.}, Pick(obj\_name), Place(obj\_name), Move(obj\_name, start\_pos, end\_pos)) based on actual appliance manipulation requirements. During inference, the model needs to select appropriate actions from these candidate atomic actions to perform planning. The evaluation uses two metrics: task completion rate and success rate. A step is correct only if its atomic action and parameters are fully accurate; the plan is correct only if all steps are correct.

\noindent\textbf{Task 3: Appliance Part Grounding}  
In the above task, the model can predict which appliance component to operate. However, for a low-level policy, the component name alone is often insufficient to determine where to perform the action. To address this, we introduce the appliance part grounding task, where the planning model uses the appliance manual and target part name to predict the its bounding box within the current appliance observation. The prediction should follow \([x_1, y_1, x_2, y_2]\) format, where \((x_1, y_1)\) and \((x_2, y_2)\) denote the top-left and bottom-right corners. This task is evaluated with average IoU and mAP@0.5.

\noindent\textbf{Task 4: Close-loop Planning Adjustment}
During appliance manipulation, external interference and execution deviations may cause discrepancies between actual outcomes and initial open‑loop plans. The planning model must therefore update plans in real time using visual feedback from appliance components (\emph{e.g.}, door state, screen display, knob position). To evaluate this, we introduce perturbations, such as opening a closed door, adjusting a knob, or changing screen content, during the manipulation. To ensure the reproducibility of the evaluation results, the locations and magnitudes of these perturbations are fixed. Given the manual, task instructions, execution history, initial planning, and real‑time observations, the model predicts the next atomic action to adjust the appliance. Closed‑loop performance is measured by step‑wise success rate.

\noindent\textbf{Task 5: Full-process Inference}
We further consider the scenario where the model executes the above tasks sequentially. Specifically, before interacting with the appliance, the model needs to perform open-loop planning based on the retrieved manual pages. Then, using the appliance component names predicted in the open-loop plan, the model locates the corresponding components in the appliance observation image. Finally, it executes the open-loop plan and adjusts the plan according to disturbances. Notably, any part localization with an IoU less than 0.5 will be considered a failure, and any incorrect action prediction will also be deemed a failure. To avoid the evaluation being affected by errors from the low-level policy, action execution is carried out through “magic manipulation.” This task is evaluated based on the plan completion rate and success rate metrics.

\subsection{Statistics about \benchmark}
To better understand the data distribution in \benchmark, we conduct statistical analysis. The benchmark contains 14 types and a total of 100 appliances and manuals. The appliance type distribution is visualized in Figure~\ref{fig:data_vis}. These appliances collectively have 589 operable components. Based on these appliances and manuals, we collected a total of 979 appliance manipulation planning tasks and 941 intermediate interference steps. The average length of our task instructions is 766.18 words, and the average length of the planning tasks is 7.57 steps. We further display the \benchmark task action transitions in Figure~\ref{fig:data_vis}. To evaluate the fidelity of our asset, we selected assets from our dataset, ArtVIP, Infinite-Mobility and Partnet-Mobility, then conducted a questionnaire with 50 participants. They rated how real the assets are on a scale of 0 to 5,  based on three aspects: size, material, and texture. The result is shown in Figure~\ref{fig:data_vis}.

\begin{table*}[tb]
    \begin{center}
    \small
    \caption{\textbf{Comparison on \benchmark independent tasks.}} 
    \scalebox{0.59}{
        \setlength{\tabcolsep}{1.0mm}{
        \begin{tabular}{r|c|c|c|c|c|c|c|c|c|c|c|c|c|c|c}
  \toprule
    \multirow{1}{*}{\centering \textbf{\textsc{Baseline Model}}}
    & \includegraphics[width=0.035\linewidth]{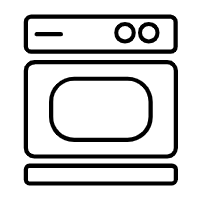}
    & \includegraphics[width=0.035\linewidth]{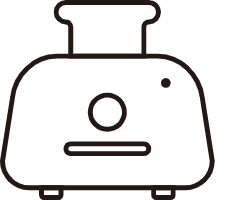}
    & \includegraphics[width=0.035\linewidth]{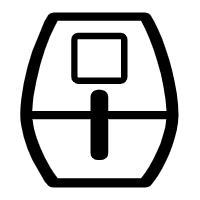}
    & \includegraphics[width=0.035\linewidth]{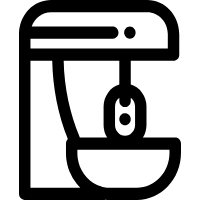}
    & \includegraphics[width=0.035\linewidth]{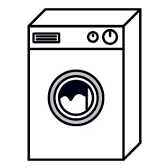}
    & \includegraphics[width=0.035\linewidth]{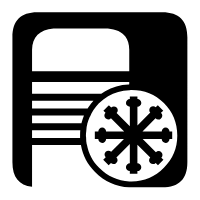}
    & \includegraphics[width=0.035\linewidth]{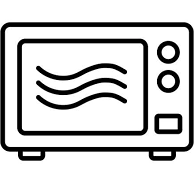}
    & \includegraphics[width=0.035\linewidth]{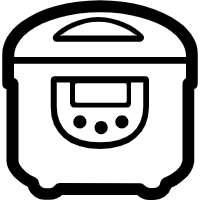}
    & \includegraphics[width=0.035\linewidth]{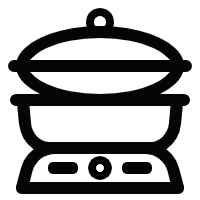}
    & \includegraphics[width=0.035\linewidth]{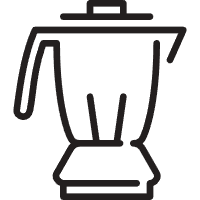}
    & \includegraphics[width=0.035\linewidth]{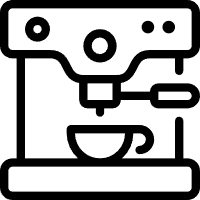}
    & \includegraphics[width=0.035\linewidth]{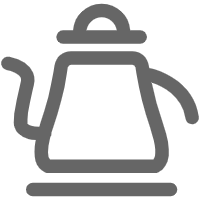}
    & \includegraphics[width=0.035\linewidth]{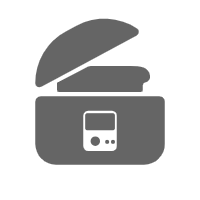}
    & \includegraphics[width=0.035\linewidth]{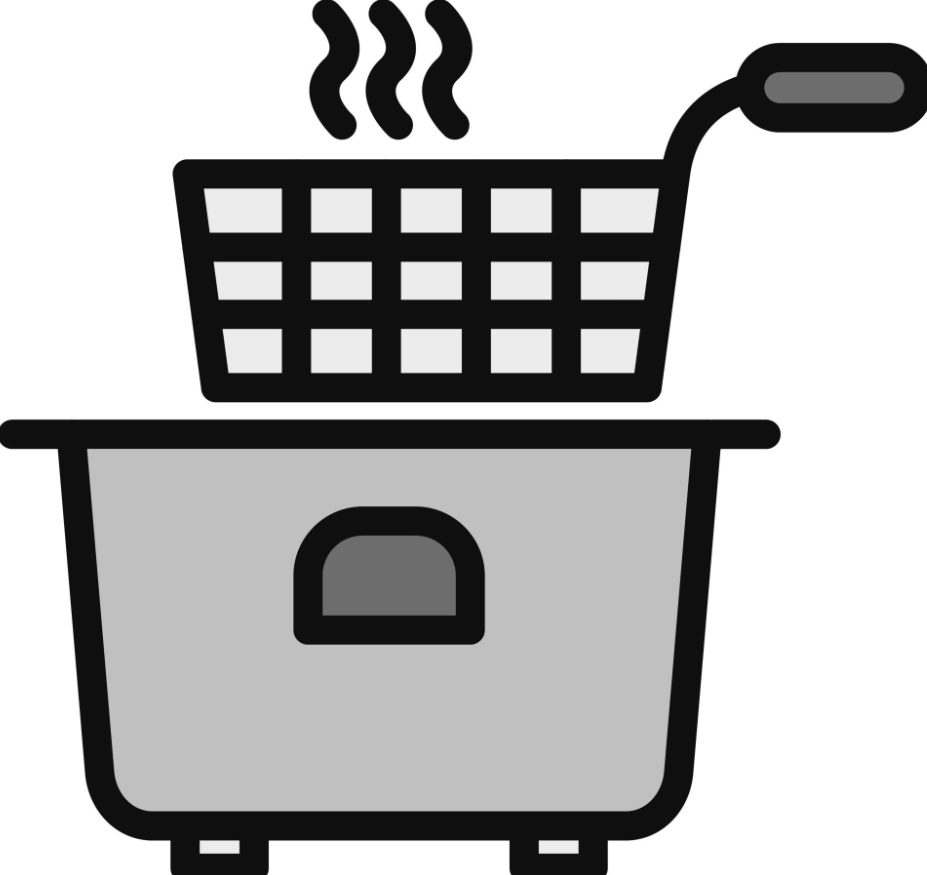}
    & \textbf{Total} \\ 
  \toprule
  \multicolumn{16}{c}{\textbf{Task 1: Manual Page Retrieval} (\emph{Recall / F1})} \\ 
  \midrule
  \rowcolor{blue!8} \multicolumn{16}{l}{\textit{Proprietary MLLMs}} \\
  GPT-5~\cite{GPT-5}  &88.88/67.53 &\textbf{75.00/70.83} &\textbf{92.30/92.30} &100.0/80.55 &90.00/86.66 &\textbf{100.0/88.88} &79.16/71.66 &100.0/88.88 &80.00/86.66 &83.33/77.77 &86.36/86.06 &50.00/58.33 &80.00/86.66 &\textbf{100.0/100.0} &86.50/80.89 \\ 
  GPT-5 Mini~\cite{GPT-5}  &83.33/57.77 &\textbf{75.00/70.83} &88.46/88.20 &100.0/78.70 &100.0/86.66 &100.0/82.22 &87.50/72.53 &100.0/77.77 &80.00/79.99 &87.50/79.20 &78.78/76.96 &\textbf{75.00/48.57} &\textbf{90.00/93.33} &\textbf{100.0/100.0} &\textbf{87.66/77.86} \\ 
  Gemini 2.5 Pro~\cite{gemini} &\textbf{94.44/68.41} &\textbf{75.00/70.83} &\textbf{92.30/92.30} &100.0/77.77 &90.00/76.00 &100.0/79.04 &\textbf{91.66/69.76} &100.0/88.88 &80.00/79.99 &\textbf{91.66/82.53} &\textbf{90.90/85.45} &50.00/45.00 &80.00/83.33 &\textbf{100.0/100.0} &90.00/79.40 \\ 
  Gemini 2.5 Flash~\cite{gemini}  &88.88/65.18 &\textbf{75.00/70.83} &88.46/89.74 &100.0/79.10 &\textbf{100.0/89.33} &100.0/82.22 &79.16/67.77 &\textbf{100.0/100.0} &80.00/86.66 &79.16/77.77 &86.36/84.54 &50.00/58.33 &80.00/86.66 &100.0/100.0 &86.00/79.85 \\ 
  \rowcolor{blue!8} \multicolumn{16}{l}{\textit{Open-source MLLMs}} \\
  Qwen3-VL 8B Thinking~\cite{Qwen3-vl}  &77.77/67.40 &68.75/64.58 &84.61/89.74 &\textbf{100.0/88.88} &70.00/74.66 &83.33/75.55 &50.000/52.50 &83.33/88.88 &70.00/73.33 &79.16/81.94 &65.15/68.18 &50.00/45.00 &70.00/79.99 &\textbf{100.0/100.0} &74.66/74.76  \\
  Qwen3-VL 8B Instruct~\cite{Qwen3-vl} &77.77/70.37 &75.00/66.66 &76.92/79.48 &100.0/80.55 &70.00/67.99 &66.66/72.22 &45.83/44.02 &83.33/83.33 &80.00/86.66 &79.16/81.94 &74.24/70.90 &50.00/40.00 &70.00/79.99 &\textbf{100.0/100.0} &74.66/72.36 \\
  Qwen3-VL 235B Thinking~\cite{Qwen3-vl}  &83.33/72.22 &62.50/62.50 &88.46/92.30 &100.0/92.59 &80.00/81.33 &100.0/82.22 &58.33/57.50 &\textbf{100.0/100.0} &80.00/86.66 &83.33/83.33 &86.36/84.54 &50.00/58.33 &70.00/79.99 &\textbf{100.0/100.0} &81.00/80.06 \\
  Qwen3-VL 235B Instruct~\cite{Qwen3-vl}  &88.88/66.53 &\textbf{75.00/70.83} &88.46/85.38 &100.0/80.55 &90.00/84.00 &100.0/79.04 &83.33/59.92 &100.0/65.55 &80.00/86.66 &83.33/75.31 &86.36/83.63 &50.00/37.50 &70.00/74.66 &\textbf{100.0/100.0} &86.00/75.78 \\
  GLM-4.1V-Thinking~\cite{glm}  &66.66/59.25 &56.25/58.33 &84.61/89.74 &\textbf{100.0/88.88} &70.00/74.66 &100.0/67.40 &54.16/54.72 &83.33/83.33 &80.00/86.66 &79.16/77.56 &78.78/83.03 &50.00/41.66 &70.00/80.00 &\textbf{100.0/100.0} &75.66/75.09 \\
  GLM-4.5V~\cite{glm}  &77.77/66.66 &62.50/62.50 &76.92/78.20 &100.0/92.59 &70.00/74.66 &66.66/66.66 &54.16/53.61 &83.33/83.33 &80.00/86.66 &79.16/80.55 &70.00/73.00 &50.00/50.00 &62.50/62.50 &\textbf{100.0/100.0} &73.97/73.94 \\
  \rowcolor{blue!8} \multicolumn{16}{l}{\textit{Embodied Planning Models}} \\
  RoboBrain 2.0-7B~\cite{robobrain2}  &0.00/0.00 &0.00/0.00 &8.33/8.33 &11.11/11.11 &0.00/0.00 &16.66/22.22 &0.00/0.00 &0.00/0.00 &10.00/13.33 &12.50/13.88 &10.00/10.00 &0.00/0.00 &0.00/0.00 &0.00/0.00 &5.78/6.31 \\
  RoboBrain 2.0-32B~\cite{robobrain2}  &72.22/64.81 &75.00/69.44 &66.66/65.00 &88.88/72.96 &40.00/46.66 &66.66/61.11 &36.36/38.48 &100.0//88.88 &80.00/61.33 &70.83/65.27 &66.66/56.16 &50.00/45.00 &70.00/70.00 &\textbf{100.0/100.0} &68.07/62.47 \\
  ManualPlan~\cite{checkmanual}  &38.88/30.00 &25.00/18.75 &66.66/68.88 &0.00/0.00 &20.00/20.00 &83.33/79.99 &50.00/44.16 &50.00/30.00 &\textbf{100.0/82.66} &62.50/44.72 &57.57/50.90 &50.00/40.00 &60.00/61.32 &66.66/50.00 &45.83/38.03 \\

  \midrule
  \multicolumn{16}{c}{\textbf{Task 2: Open-loop Manipulation Planning} (\emph{Task Completion Rate / Task Success Rate})} \\ 
  \midrule
  \rowcolor{blue!8} \multicolumn{16}{l}{\textit{Proprietary MLLMs}} \\ 
  GPT-5~\cite{GPT-5}  &2.60/1.11 &2.57/0.00 &1.55/0.00 &0.00/0.00 &7.77/2.00 &5.92/0.00 &4.70/0.83 &0.95/0.00 &4.80/2.00 &3.02/0.00 &6.44/2.02 &7.77/0.00 &14.59/6.00 &10.66/10.00 &4.30/1.22  \\ 
  GPT-5 Mini~\cite{GPT-5}  &2.22/2.22 &5.23/1.25 &0.11/0.00 &0.00/0.00 &3.58/0.00 &1.26/0.00 &3.52/1.66 &6.45/0.00 &4.09/2.00 &2.34/0.00 &4.08/0.00 &5.55/0.00 &9.29/4.00 &9.66/6.66 &3.27/1.02 \\ 
  Gemini 2.5 Pro~\cite{gemini}  &0.59/0.00 &3.47/1.25 &3.70/1.53 &0.00/0.00 &3.18/0.00 &2.01/0.00 &4.51/4.16 &0.47/0.00 &0.50/0.00 &1.07/0.00 &\textbf{10.72/8.08} &8.88/0.00 &\textbf{15.80/12.00} &8.00/6.66 &4.08/2.45 \\ 
  Gemini 2.5 Flash~\cite{gemini}  &0.23/0.00 &2.00/0.00 &\textbf{6.12/3.84} &0.00/0.00 &2.79/0.00 &1.26/0.00 &2.34/1.66 &7.82/3.33 &2.16/2.00 &2.34/0.88 &10.55/8.08 &5.55/0.00 &15.51/12.00 &6.66/6.66 &4.26/2.65 \\ 
  \rowcolor{blue!8} \multicolumn{16}{l}{\textit{Open-source MLLMs}} \\
  Qwen3-VL 8B Thinking~\cite{Qwen3-vl}  &1.55/1.11 &2.82/0.00 &1.53/1.53 &0.00/0.00 &3.88/2.00 &4.23/3.70 &1.79/1.66 &7.42/3.33 &5.73/2.00 &0.88/0.88 &6.39/5.05 &2.77/0.00 &7.22/4.00 &6.66/6.66 &3.01/1.94 \\
  Qwen3-VL 8B Instruct~\cite{Qwen3-vl}  &0.00/0.00 &4.41/0.00 &0.76/0.76 &0.00/0.00 &0.00/0.00 &0.00/0.00 &1.66/1.66 &0.00/0.00 &2.00/2.00 &0.88/0.88 &1.68/0.00 &\textbf{12.22/0.00} &6.71/4.65 &3.33/3.33 &1.70/0.82 \\
  Qwen3-VL 235B Thinking~\cite{Qwen3-vl}  &1.56/1.11 &5.95/1.25 &1.59/0.00 &0.00/0.00 &\textbf{9.46/4.00} &7.83/3.70 &2.08/1.66 &\textbf{10.79/3.33 }&5.40/2.00 &4.18/0.88 &6.68/4.04 &8.33/0.00 &7.51/2.00 &8.00/6.66 &4.36/1.73 \\
  Qwen3-VL 235B Instruct~\cite{Qwen3-vl}  &0.00/0.00 &6.86/1.25 &3.23/3.07 &0.00/0.00 &8.01/4.00 &2.32/0.00 &4.50/4.16 &3.14/0.00 &3.13/2.00 &2.47/1.76 &6.58/3.03 &8.88/0.00 &9.95/6.00 &6.66/6.66 &4.11/2.34 \\
  GLM-4.1V-Thinking~\cite{glm}  &0.00/0.00 &0.00/0.00 &0.00/0.00 &0.00/0.00 &0.00/0.00 &0.00/0.00 &0.00/0.00 &0.00/0.00 &0.00/0.00 &0.00/0.00 &0.00/0.00 &0.00/0.00 &0.00/0.00 &0.00/0.00 &0.00/0.00 \\
  GLM-4.5V~\cite{glm}  &2.22/2.22 &3.30/1.25 &2.41/2.30 &\textbf{1.11/1.11} &5.21/4.00 &4.44/3.70 &1.30/0.83 &1.33/0.00 &4.43/2.00 &1.50/0.97 &8.41/6.06 &3.33/0.00 &10.11/8.00 &\textbf{12.66/10.00} &3.73/2.68 \\
  \rowcolor{blue!8} \multicolumn{16}{l}{\textit{Embodied Planning Models}} \\
  RoboBrain 2.0-7B~\cite{robobrain2}  &0.00/0.00 &0.84/0.00 &0.06/0.00 &0.00/0.00 &0.17/0.00 &0.00/0.00 &0.16/0.00 &0.00/0.00 &0.00/0.00 &0.27/0.00 &0.00/0.00 &0.55/0.00 &0.22/0.00 &0.00/0.00 &0.16/0.00 \\
  RoboBrain 2.0-32B~\cite{robobrain2}  &0.00/0.00 &1.99/0.00 &0.00/0.00 &0.00/0.00 &0.71/0.00 &0.00/0.00 &0.12/0.00 &0.00/0.00 &0.00/0.00 &0.17/0.00 &0.67/0.00 &0.00/0.00 &1.53/0.00 &0.00/0.00 &0.37/0.00 \\
  ManualPlan~\cite{checkmanual}  &\textbf{6.12/0.00} &\textbf{8.76/0.00} &2.003/0.00 &0.92/0.00 &3.20/0.00 &5.75/0.00 &\textbf{11.27/0.00 }&10.44/0.00 &\textbf{6.44/0.00} &4.51/1.76 &2.20/0.00 &11.32/0.00 &11.50//3.99 &2.17/0.00 &\textbf{5.61/0.40} \\
  ApBot~\cite{apbot}  &0.00/0.00 &0.00/0.00 &0.00/0.00 &0.00/0.00 &2.00/2.00 &\textbf{8.70/7.40} &1.60/0.80 &0.00/0.00 &0.00/0.00 &\textbf{7.10/7.10} &7.10/7.10 &0.00/0.00 &4.40/4.00 &0.00/0.00 &2.30/2.10\\
  \midrule
  \multicolumn{16}{c}{\textbf{Task 3: Appliance Part Grounding} (\emph{Average / mAP@0.5})} \\
  \midrule
  \rowcolor{blue!8} \multicolumn{16}{l}{\textit{Proprietary MLLMs}} \\
  GPT-5~\cite{GPT-5}  &12.71/7.69 &4.84/0.00 &\textbf{12.00/8.51} &\textbf{16.33/17.85} &13.23/9.67 &\textbf{28.83/28.57} &\textbf{20.24/11.94} &\textbf{10.42/4.34} &\textbf{32.97/33.33} &10.01/9.67 &4.72/1.33 &\textbf{19.17/18.75} &11.17/9.67 &4.80/0.00 &\textbf{12.15/8.59} \\ 
  GPT-5 Mini~\cite{GPT-5}  &9.08/7.46 &1.85/0.00 &7.18/3.19 &9.71/0.00 &4.43/3.22 &10.83/0.00 &13.95/14.92 &2.45/0.00 &19.56/0.00 &1.82/0.00 &2.48/0.00 &7.59/6.25 &5.33/0.00 &4.94/0.00 &6.51/3.49 \\ 
  Gemini 2.5 Pro~\cite{gemini}  &7.77/4.47 &3.75/0.00 &10.74/8.51 &9.47/10.71 &6.74/6.45 &7.23/0.00 &10.46/8.95 &5.87/4.34 &32.82/41.66 &9.70/11.29 &2.92/0.00 &6.64/6.25 &6.70/6.45 &6.16/0.00 &8.16/6.64 \\ 
  Gemini 2.5 Flash~\cite{gemini}  &8.21/4.47 &2.46/0.00 &7.94/7.44 &12.14/17.85 &2.28/3.22 &16.67/0.00 &11.22/7.46 &2.18/0.00 &21.16/16.66 &6.06/4.83 &1.52/1.33 &5.06/0.00 &4.81/6.45 &8.38/0.00 &6.67/5.06 \\ 
  \rowcolor{blue!8} \multicolumn{16}{l}{\textit{Open-source MLLMs}} \\
  Qwen3-VL 8B Thinking~\cite{Qwen3-vl}  &5.55/1.49 &0.25/0.00 &2.14/1.06 &2.02/0.00 &1.11/0.00 &5.29/0.00 &6.09/0.00 &3.93/0.00 &5.25/0.00 &2.52/1.61 &0.40/0.00 &0.00/0.00 &3.03/3.22 &9.93/0.00 &2.92/0.69 \\
  Qwen3-VL 8B Instruct~\cite{Qwen3-vl}  &2.27/0.00 &0.00/0.00 &0.44/0.00 &0.63/0.00 &1.28/0.00 &0.15/0.00 &2.83/0.00 &1.56/0.00 &4.58/0.00 &0.85/0.00 &0.32/0.00 &0.00/0.00 &1.46/0.00 &8.32/0.00 &1.32/0.00 \\
  Qwen3-VL 235B Thinking~\cite{Qwen3-vl}  &5.19/2.98 &0.99/0.00 &3.41/3.19 &1.01/0.00 &1.96/0.00 &0.25/0.00 &6.78/0.00 &1.58/0.00 &6.01/0.00 &1.08/0.00 &0.12/0.00 &0.00/0.00 &1.50/0.00 &8.64/0.00 &2.80/0.87 \\
  Qwen3-VL 235B Instruct~\cite{Qwen3-vl}  &4.58/4.47 &0.00/0.00 &0.61/0.00 &0.61/0.00 &1.15/0.00 &0.46/0.00 &4.64/0.00 &1.61/0.00 &6.28/0.00 &0.86/0.00 &0.00/0.00 &0.00/0.00 &0.44/0.00 &7.35/0.00 &1.75/0.52 \\
  GLM-4.1V-Thinking~\cite{glm}  &3.88/1.49 &0.00/0.00 &0.54/0.00 &0.98/0.00 &1.67/0.00 &0.42/0.00 &4.01/0.00 &1.55/0.00 &4.71/0.00 &0.93/0.00 &0.00/0.00 &0.00/0.00 &1.33/0.00 &\textbf{11.63/0.00} &1.74/0.17 \\
  GLM-4.5V~\cite{glm}  &3.59/2.98 &0.00/0.00 &0.75/0.00 &0.99/0.00 &2.50/0.00 &0.32/0.00 &4.09/0.00 &1.56/0.00 &6.15/0.00 &0.66/0.00 &0.00/0.00 &0.00/0.00 &0.62/0.00 &11.00/0.00 &1.76/0.35 \\
  \rowcolor{blue!8} \multicolumn{16}{l}{\textit{Embodied Planning Models}} \\
  RoboBrain 2.0-7B~\cite{robobrain2}  &0.00/0.00 &0.00/0.00 &0.00/0.00 &0.00/0.00 &0.00/0.00 &0.00/0.00 &0.00/0.00 &0.00/0.00 &0.00/0.00 &0.00/0.00 &0.00/0.00 &0.00/0.00 &0.00/0.00 &0.00/0.00 &0.00/0.00 \\
  RoboBrain 2.0-32B~\cite{robobrain2}  &0.00/0.00 &0.00/0.00 &0.00/0.00 &0.00/0.00 &0.00/0.00 &0.00/0.00 &0.00/0.00 &0.00/0.00 &0.00/0.00 &0.00/0.00 &0.00/0.00 &0.00/0.00 &0.00/0.00 &0.00/0.00 &0.00/0.00 \\
  ManualPlan~\cite{checkmanual}  &3.56/0.00 &0.00/0.00 &1.01/0.00 &2.46/0.00 &2.37/0.00 &3.08/0.00 &5.61/0.00 &1.25/0.00 &2.79/0.00 &0.20/0.00 &0.15/0.00 &1.66/0.00 &1.73/0.00 &5.07/0.00 &1.92/0.00 \\
  ApBot~\cite{apbot}  &\textbf{23.60/28.40} &\textbf{10.50/12.30} &8.50/7.60 &4.70/4.50 &\textbf{15.70/20.00} &17.70/20.00 &10.20/11.20 &0.00/0.00 &4.50/4.30 &\textbf{11.70/12.00} &\textbf{11.20/14.70} &0.00/0.00 &\textbf{12.20/13.60} &4.30/5.30 &10.60/12.10 \\
  \midrule
  \multicolumn{16}{c}{\textbf{Task 4: Close-loop Planning Adjustment} (\emph{Step-wise Success Rate})} \\
  \midrule
  \rowcolor{blue!8} \multicolumn{16}{l}{\textit{Proprietary MLLMs}} \\ 
  GPT-5~\cite{GPT-5}  &28.84 &5.79 &\textbf{38.51} &0.00 &31.74 &12.50 &39.43 &6.66 &\textbf{41.66} &30.55 &20.58 &10.52 &34.14 &\textbf{45.45} &29.61 \\ 
  GPT-5 Mini~\cite{GPT-5}  &14.42 &5.17 &18.24 &0.00 &15.78 &3.84 &25.19 &15.00 &37.83 &9.09 &20.58 &42.10 &0.00 &27.27 &16.33 \\ 
  Gemini 2.5 Pro~\cite{gemini}  &26.13 &13.04 &45.62 &0.00 &42.85 &11.53 &38.00 &5.00 &27.02 &35.45 &23.52 &15.78 &\textbf{41.46} &31.81 &31.73 \\ 
  Gemini 2.5 Flash~\cite{gemini}  &\textbf{31.06} &10.14 &34.05 &0.00 &46.03 &19.23 &37.90 &15.00 &24.32 &37.03 &26.47 &\textbf{42.10} &39.02 &33.33 &31.61 \\ 
  \rowcolor{blue!8} \multicolumn{16}{l}{\textit{Open-source MLLMs}} \\
  Qwen3-VL 8B Thinking~\cite{Qwen3-vl}  &23.07 &13.04 &30.43 &0.00 &30.15 &11.53 &36.36 &15.00 &13.51 &37.27 &20.58 &36.84 &0.00 &22.72 &25.58 \\
  Qwen3-VL 8B Instruct~\cite{Qwen3-vl}  &28.84 &24.63 &36.95 &0.00 &38.09 &19.23 &\textbf{39.86} &10.00 &8.10 &38.18 &23.52 &\textbf{42.10} &26.86 &4.54 &30.65 \\
  Qwen3-VL 235B Thinking~\cite{Qwen3-vl}  &22.11 &14.49 &33.33 &0.00 &41.26 &23.07 &38.56 &15.00 &35.13 &38.18 &\textbf{32.35} &\textbf{42.10} &29.26 &40.90 &31.23 \\
  Qwen3-VL 235B Instruct~\cite{Qwen3-vl}  &25.00 &24.63 &39.13 &0.00 &42.85 &26.92 &35.94 &10.00 &5.40 &28.18 &26.47 &\textbf{42.10} &24.39 &9.09 &29.13 \\
  GLM-4.1V-Thinking~\cite{glm}  &0.00 &0.00 &0.00 &0.00 &0.00 &0.00 &0.00 &0.00 &0.00 &0.00 &0.00 &0.00 &0.00 &0.00 &0.00 \\
  GLM-4.5V~\cite{glm}  &0.00 &0.00 &0.74 &0.00 &0.00 &0.00 &0.00 &0.00 &0.00 &0.00 &0.00 &0.00 &0.00 &0.00 &0.12 \\
  \rowcolor{blue!8} \multicolumn{16}{l}{\textit{Embodied Planning Models}} \\
  RoboBrain 2.0-7B~\cite{robobrain2}  &27.27 &\textbf{25.67} &34.31 &0.00 &\textbf{45.16} &\textbf{53.33} &34.86 &\textbf{22.72} &5.26 &\textbf{40.77} &27.77 &36.84 &37.03 &9.09 &\textbf{31.77} \\
  RoboBrain 2.0-32B~\cite{robobrain2}  &23.14 &20.27 &28.99 &0.00 &22.58 &13.33 &26.31 &18.18 &7.89 &18.44 &19.44 &26.31 &25.92 &9.09 &21.96 \\
  ApBot~\cite{apbot}  &0.80 &2.60 &4.40 &0.00 &18.20 &25.00 &3.70 &2.80 &5.80 &9.90 &16.70 &26.30 &5.90 &0.00 &7.00 \\
  \midrule
  \multicolumn{16}{c}{\textbf{Task 5: Full-process Inference task} (\emph{Task Completion Rate / Task Success Rate})} \\
  \midrule
  \rowcolor{blue!8} \textit{Proprietary MLLMs} & & & & & & & & & & & & & & & \\ 
  GPT-5~\cite{GPT-5}  &0.00/0.00  &\textbf{2.32/0.00}  &0.00/0.00  &0.00/0.00  &0.00/0.00  &0.00/0.00  &0.00/0.00  &0.00/0.00  &0.00/0.00  &0.19/0.00  &1.99/0.00  &0.55/0.00  &0.00/0.00  &0.00/0.00  &0.44/0.00  \\ 
  GPT-5 Mini~\cite{GPT-5}  &0.00/0.00 &1.27/0.00 &0.00/0.00 &0.00/0.00 &0.00/0.00 &0.00/0.00 &0.00/0.00 &0.00/0.00 &0.00/0.00 &0.44/0.00 &1.89/0.00 &2.77/0.00 &0.22/0.00 &0.00/0.00 &0.41/0.00 \\
  Gemini 2.5 Pro~\cite{gemini}  &\textbf{0.44/0.00} &2.22/0.00 &0.00/0.00 &0.00/0.00 &\textbf{0.59/0.00} &0.00/0.00 &0.06/0.00 &0.00/0.00 &0.00/0.00 &0.44/0.00 &2.47/0.00 &\textbf{8.88/0.00} &0.40/0.00 &0.00/0.00 &0.77/0.00 \\
  Gemini 2.5 Flash~\cite{gemini}  &0.00/0.00 &0.35/0.00 &0.00/0.00 &0.00/0.00 &0.00/0.00 &0.00/0.00 &0.00/0.00 &0.00/0.00 &0.20/0.00 &0.61/0.00 &1.88/0.00 &3.17/0.00 &\textbf{1.29/0.00} &0.00/0.00 &0.41/0.00 \\
  \rowcolor{blue!8} \textit{Open-source MLLMs} & & & & & & & & & & & & & & & \\
  Qwen3-VL 235B Thinking~\cite{Qwen3-vl}  &0.00/0.00 &1.13/0.00 &0.00/0.00 &0.00/0.00 &0.00/0.00 &0.00/0.00 &0.00/0.00 &0.00/0.00 &0.00/0.00 &0.11/0.00 &0.45/0.00 &2.22/0.00 &0.00/0.00 &0.00/0.00 &0.20/0.00 \\
  Qwen3-VL 235B Instruct~\cite{Qwen3-vl}  &0.00/0.00 &0.75/0.00 &0.00/0.00 &0.00/0.00 &0.00/0.00 &0.00/0.00 &0.00/0.00 &0.00/0.00 &0.00/0.00 &0.00/0.00 &0.00/0.00 &0.55/0.00 &0.00/0.00 &0.00/0.00 &0.07/0.00 \\
  Qwen3-VL 8B Thinking~\cite{Qwen3-vl}  &0.00/0.00 &0.55/0.00 &0.00/0.00 &0.00/0.00 &0.00/0.00 &0.00/0.00 &0.00/0.00 &0.00/0.00 &\textbf{0.33/0.00} &0.00/0.00 &0.00/0.00 &1.66/0.00 &0.00/0.00 &0.00/0.00 &0.09/0.00 \\
  Qwen3-VL 8B Instruct~\cite{Qwen3-vl}  &0.00/0.00 &1.19/0.00 &0.00/0.00 &0.00/0.00 &0.00/0.00 &0.00/0.00 &0.00/0.00 &0.00/0.00 &0.00/0.00 &0.00/0.00 &0.00/0.00 &0.00/0.00 &0.00/0.00 &0.00/0.00 &0.09/0.00    \\
  GLM-4.1V-Thinking~\cite{glm}  &0.00/0.00 &0.00/0.00 &0.00/0.00 &0.00/0.00 &0.00/0.00 &0.00/0.00 &0.00/0.00 &0.00/0.00 &0.00/0.00 &0.00/0.00 &0.00/0.00 &0.00/0.00 &0.00/0.00 &0.00/0.00 &0.00/0.00 \\
  GLM-4.5V~\cite{glm}  &0.00/0.00 &0.00/0.00 &0.00/0.00 &0.00/0.00 &0.00/0.00 &0.00/0.00 &0.00/0.00 &0.00/0.00 &0.00/0.00 &0.00/0.00 &0.00/0.00 &0.00/0.00 &0.00/0.00 &0.00/0.00 &0.00/0.00 \\
  \rowcolor{blue!8} \textit{Embodied Planning Models} & & & & & & & & & & & & & & & \\
  RoboBrain 2.0-7B~\cite{robobrain2}  &0.00/0.00 &0.00/0.00 &0.00/0.00 &0.00/0.00 &0.00/0.00 &0.00/0.00 &0.00/0.00 &0.00/0.00 &0.00/0.00 &0.00/0.00 &0.00/0.00 &0.00/0.00 &0.00/0.00 &0.00/0.00 &0.00/0.00    \\
  RoboBrain 2.0-32B~\cite{robobrain2}  &0.00/0.00 &0.00/0.00 &0.00/0.00 &0.00/0.00 &0.00/0.00 &0.00/0.00 &0.00/0.00 &0.00/0.00 &0.00/0.00 &0.00/0.00 &0.00/0.00 &0.00/0.00 &0.00/0.00 &0.00/0.00 &0.00/0.00    \\
  ApBot~\cite{apbot}  &0.00/0.00 &0.00/0.00 &0.00/0.00 &0.00/0.00 &0.00/0.00 &\textbf{7.40/7.40 }&\textbf{0.83/0.83} &0.00/0.00 &0.00/0.00 &\textbf{1.77/1.77} &\textbf{4.04/4.04} &0.00/0.00 &0.00/0.00 &0.00/0.00 &\textbf{0.92/0.92}  \\
  \bottomrule
        \end{tabular}}
    }
    \vspace{-0.3cm}
 \label{tab:single_task}    
    \vspace{-0.3cm}
    \end{center}
\end{table*}

\section{Evaluation}
\subsection{Baseline Models}
\begin{itemize}
    \item[$\bullet$] \textbf{Proprietary MLLMs}: We chose the latest GPT models recently released by OpenAI, including GPT-5 and GPT-5 Mini~\cite{GPT-5}, as well as Google's latest models, Gemini 2.5 Pro and Gemini 2.5 Flash~\cite{gemini}.
    \item[$\bullet$] \textbf{Open-source MLLMs}: We selected Qwen3-VL~\cite{Qwen3-vl} models of different sizes for evaluation, including the 235B Thinking model and Instruct model, as well as the 8B Thinking model and Instruct model. We also evaluate GLM-4.5V 106B and GLM-4.1V-Thinking 9B~\cite{glm} models.
    \item[$\bullet$] \textbf{Embodied Planning Models}: We selected Robobrain 2.0 7B and 32B~\cite{robobrain2}, which claim to have generalization capabilities in embodied manipulation planning, as well as the planning models ManualPlan~\cite{checkmanual} and Apbot~\cite{apbot} oriented toward appliance manipulation.
\end{itemize}

\begin{figure*}[htp]
    \centering
    \includegraphics[width=0.9\linewidth]{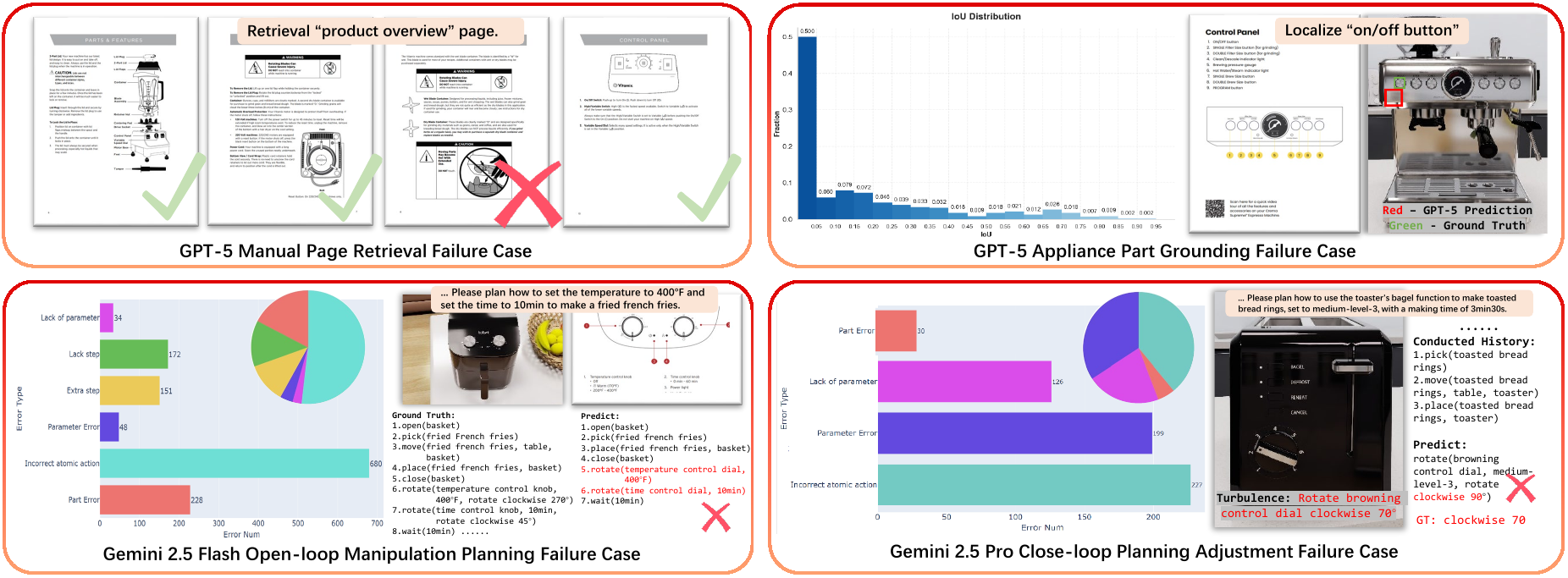}
    \vspace{-0.3cm}
    \caption{\textbf{Failure case study on \benchmark.}}
    \label{fig:case_study}
    \vspace{-0.3cm}
\end{figure*}

\subsection{Performance on \benchmark}
\noindent\textbf{How lagre models perform on manual page retrieval?}
From Table~\ref{tab:single_task}, we can observe that proprietary multimodal large models outperform open-source multimodal large models, which in turn outperform end-to-end embodied planning models. This phenomenon reflects that during the training of the embodied planning model like robobrain, due to the lack of document understanding data, the model's document understanding exhibits a certain degree of forgetting as training progresses. This forgetting may affect the manual-based manipulation planning. Increasing training data related to document understanding for home appliance operation might help maintain the basic document understanding ability within the embodied planning model.

\noindent\textbf{How is the quality of the appliance manipulation plan predicted by the large model?}
From Table~\ref{tab:single_task}, manual-based manipulation planning for appliances remains highly challenging for current models. Figure~\ref{fig:case_study} visualizes predictions from Gemini 2.5 Flash, one of the best-performing model, highlighting three major failure types: Incorrect atomic action, Part error, and Missing steps. Incorrect atomic action (\emph{e.g.}, using Set instead of Rotate for a knob) shows the model’s confusion over suitable operations for different components. Part error reflects limited understanding of manuals and part descriptions. Missing steps indicate insufficient appliance logic comprehension leading to incomplete sequences. To improve the situation, large model may need more powerful reasoning ability about appliance operation logic.

\noindent\textbf{Can large models correctly identify appliance components based on the manuals?}
The evaluation results about appliance part grounding is shown in Table~\ref{tab:single_task}. Existing models struggle to predict effective bounding boxes for these parts. To analyze prediction errors, we visualize the IoU distribution from GPT-5 in Figure~\ref{fig:case_study}, where most IoUs fall in [0, 0.05]. Despite large models' strong object detection capabilities, challenges stem from aligning manual images to real-time observations. Their limited spatial intelligence~\cite{1983Frames} make accurate localization difficult when viewing angles differ from the manual. Therefore, exploring ways to enhance the model's ability to associate multimodal information and its spatial reasoning skills could be valuable for the research.

\noindent\textbf{How can model timely adjust initial plan to adapt appliance status change?}
We can observe the close-loop planning adjustment task results in Table~\ref{tab:single_task}. From the table, large model have a limited capability to adapt real-time changes. We visualized the failure cases in Figure~\ref{fig:case_study}. We find Incorrect atomic action still being the most serious failure reason, showing the large model failed to sense the change on appliance status and adjust the planning correctly. Secondly, Parameter error ranked second place in the table. This indicates the large model lack fine-grained vision perception on the manual and observation image, cannot accurately detect how different the current status and the target status of the appliance is. A possible approach is enhancing the model's capacity to comprehend fine-grained visual elements and improving its multimodal reasoning capabilities. 

\noindent\textbf{Can existing models complete the full-process appliance manipulation planning?}
Almost all model completely fail to complete even one full-process task, which makes existing models all achieve zero success rates in Table~\ref{tab:single_task}. Their low performance comes from the error accumulation during the full-process inference. Therefore, it is necessary for the large model to perform better on single tasks to enhancing its full-process appliance operation capabilities





\section{Limitations and Future Work}
In this paper, we focus on how appliance assets aligned with real-world instruction manuals can facilitate the evaluation of appliance operation planning. While our study centers on the planning level, these digital assets also hold significant potential for advancing low-level appliance operation policies — an avenue yet to be explored in our current work. For future research, RealAppliance could be leveraged to collect large-scale appliance operation data for training vision-language-action models, or to establish standardized benchmarks for low-level manipulation.

{
    \small
    \bibliographystyle{ieeenat_fullname}
    \bibliography{main}
}

\end{document}